\documentclass{article}

\usepackage{microtype}
\usepackage{graphicx}
\usepackage{subcaption}
\usepackage{booktabs} 

\usepackage{hyperref}

\usepackage[preprint]{icml2026}

\usepackage{amsmath}
\usepackage{amssymb}
\usepackage{mathtools}
\usepackage{amsthm}

\usepackage[capitalize,noabbrev]{cleveref}

\theoremstyle{plain}
\newtheorem{theorem}{Theorem}[section]

\newtheorem{lemma}[theorem]{Lemma}
\newtheorem{corollary}[theorem]{Corollary}
\theoremstyle{definition}

\theoremstyle{remark}

\usepackage{algorithm}

\usepackage{algpseudocode}

\usepackage{tikz}
\usetikzlibrary{arrows.meta, positioning, shapes.geometric}

\usepackage{enumitem}

\newcommand{\method}{\textsc{REDEREF}}
\newcommand{\BetaD}{\mathrm{Beta}}

\usepackage[disable,textsize=tiny]{todonotes}

\icmltitlerunning{Training-Free Agentic AI for Multi-Agent LLM Systems}

\begin{document}

\twocolumn[
  \icmltitle{Training-Free Agentic AI: Probabilistic Control and Coordination in Multi-Agent LLM Systems}

  \icmlsetsymbol{equal}{*}

\begin{icmlauthorlist}
  \icmlauthor{Mohammad Parsa Hosseini}{equal,acc}
  \icmlauthor{Ankit Shah}{equal,acc}
  \icmlauthor{Saiyra Qureshi}{acc}
  \icmlauthor{Alex Huang}{acc}
  \icmlauthor{Connie Miao}{acc}
  \icmlauthor{Wei Wei}{acc}
\end{icmlauthorlist}

\icmlaffiliation{acc}{Accenture}

\icmlcorrespondingauthor{Parsa Hosseini}{parsa.hosseini@accenture.com}
\icmlcorrespondingauthor{Ankit Shah}{ankit.parag.shah@accenture.com}

\icmlkeywords{Multi-agent systems, large language models, routing, efficiency, Thompson sampling}

  \vskip 0.3in
]

\printAffiliationsAndNotice{\icmlEqualContribution}

\begin{abstract}
Multi-agent large language model (LLM) systems enable complex, long-horizon reasoning by composing specialized agents, but practical deployment remains hindered by inefficient routing, noisy feedback,
and high interaction cost. We introduce \textsc{REDEREF}, a lightweight and training-free controller
for multi-agent LLM collaboration that improves \emph{routing efficiency during recursive delegation}.
\textsc{REDEREF} integrates (i) belief-guided delegation via Thompson sampling to prioritize agents
with historically positive marginal contributions, (ii) reflection-driven re-routing using a calibrated
LLM or programmatic judge, (iii) evidence-based selection rather than output averaging, and
(iv) memory-aware priors to reduce cold-start inefficiency.
Across multi-agent split-knowledge tasks, we show that while recursive retry alone saturates task
success, belief-guided routing reduces token usage by 28\%, agent calls by 17\%, and time-to-success by 19\% compared to random recursive delegation,
and adapts gracefully under agent or judge degradation. These results demonstrate that simple,
interpretable probabilistic control can meaningfully improve the efficiency and robustness of multi-agent LLM systems without training or fine-tuning.
\end{abstract}

\section{Introduction}

Large Language Models (LLMs) such as GPT-4, Gemini, and Claude have evolved from narrow text-completion tools into broadly capable reasoning engines, surpassing human performance on tasks ranging from legal exams~\cite{openai2023gpt4} to software engineering challenges~\cite{swebench2023}. However, most of these benchmarks are \textit{single-turn} interactions. Real-world deployments, in contrast, demand persistence and adaptability: evolving software repositories over weeks, synthesizing iterative scientific reviews, or mediating multistakeholder corporate decisions. These are inherently multiturn, long-horizon, and collaborative, pushing beyond what any single monolithic model, no matter how large, can reliably achieve.

Multi-agent LLM systems have emerged as a promising direction to scale intelligence by composing the complementary skills of multiple agents. For example, one agent might specialize in writing unit tests, another in literature surveys, and a third in policy analysis. By enabling interaction between such experts, these systems can address tasks outside the scope of any individual model. Yet, their deployment remains hampered by three persistent bottlenecks:

\textbf{(1) Dynamic task routing.} Existing orchestrators often rely on fixed pipelines or static vector similarity rules. These approaches are brittle: When task requirements change or agent performance degrades, the system continues to misroute, invoking wrong experts, and compounding errors.

\textbf{(2) Credit assignment across long horizons.} In extended dialogues, failures may not be visible until dozens of turns later. Without timely and fine-grained feedback, the system cannot effectively demote underperforming agents or up-weight reliable ones, leading to stagnation in routing efficiency.

\textbf{(3) Cold-start inefficiency.} When a new task arrives, the system has no prior evidence about which agents are competent. Without mechanisms to transfer knowledge from previous interactions, early routing is effectively random, wasting tokens and agent calls before the system can differentiate agent quality.

We propose \method, a lightweight and training-free controller for multi-agent LLM systems.
\method\ wraps any pool of agents with four components:
(1) belief-guided delegation via Thompson sampling to prioritize agents with historically positive marginal contributions;
(2) calibrated reflection via a judge that triggers credit assignment and re-routing;
(3) evidence-checked selection (rather than averaging); and
(4) memory-aware priors to reduce cold-start inefficiency and context bloat.

By combining these elements into a recursive loop, \method\ enables efficient, belief-guided routing across a pool of heterogeneous agents. In controlled experiments on split-knowledge tasks, we show that belief-guided routing reduces token usage by 28\%, agent calls by 17\%, and time-to-success by 19\% compared to an otherwise identical system with uniform random selection, while maintaining matched task success rates.

\paragraph{Contributions.}
We make the following contributions:
\begin{itemize}
    \item We introduce \textsc{REDEREF}, a training-free controller for multi-agent LLM systems that
    improves routing efficiency under recursive retry without requiring fine-tuning or centralized training.
    \item We reinterpret belief updates as modeling an agent's probability of providing a
    \emph{positive marginal contribution}, resolving credit assignment in compositional multi-agent tasks.
    \item We demonstrate empirically that belief-guided delegation substantially reduces token usage,
    agent calls, and time-to-success compared to random recursive delegation at matched success rates.
    \item We analyze robustness to agent impairment and judge miscalibration, showing graceful
    degradation rather than catastrophic routing failures.
\end{itemize}

\section{Related Work}
\label{sec:related}

\paragraph{Reasoning, reflection, and search.}
Methods such as ReAct~\cite{react2023}, Self-Refine~\cite{selfrefine2023}, Reflexion~\cite{reflexion2023}, and Tree-of-Thoughts (ToT)~\cite{tot2023} improve reasoning by coupling tool use, iterative self-feedback, episodic memory, or backtracking search. Agent-R~\cite{agentr2025} and recent surveys~\cite{guo2024survey} highlight reflection as central to multi-agent LLMs. Unlike these approaches, \method\ uses an \emph{explicit probabilistic controller}: Thompson sampling delegates under uncertainty, while binary judge outcomes update interpretable Beta posteriors that drive re-routing.

\paragraph{Orchestration, ensembles, and routing.}
Frameworks such as AutoGen~\cite{autogen2023} coordinate agents via scripted protocols, and Mixture-of-Agents (MoA)~\cite{moa2024} aggregate outputs via ensembles. Early orchestrators (e.g., LangChain graphs, AutoGPT workflows) hard-code pipelines, while systems like RopMura~\cite{ropmura2025}, DyLAN~\cite{dylan2023}, and MLPO~\cite{mlpo2025} introduce dynamic routing with agent team optimization. In contrast, \method\ performs online belief-guided delegation without retraining, maintaining efficiency and decentralization.

\paragraph{Learning-based coordination.}
Multi-agent reinforcement learning has long addressed coordination~\cite{lowe2017multi,foerster2018counterfactual}, with adaptations for LLMs such as SWEET-RL~\cite{zhou2025sweetrl}. Yet RL methods are sample-hungry. Probabilistic approaches, e.g., Bayesian Delegation~\cite{Wu21}, estimate expertise from sparse data. \method\ combines Thompson sampling with reflection-driven control,
cooldown-based exploration, and memory-aware priors in a unified,
training-free orchestration loop.

\paragraph{Benchmarks and positioning.}
Public environments (WebArena~\cite{webarena2023}, Mind2Web~\cite{mind2web2023}, GAIA~\cite{gaia2023}, SWE-bench~\cite{swebench2023}) highlight the difficulty of real-world, long-horizon tasks (e.g., GPT-4 achieves $\sim$14\% on WebArena vs.\ humans at $\sim$78\%). Against this backdrop, \method\ contributes a training-free adaptive controller that combines belief-guided delegation, embedded reflection and memory to improve efficiency, robustness and interpretability over static, ensemble or RL-based approaches.

\section{The \method\ Framework}
\label{sec:method}

\begin{figure*}[t]
\centering
\resizebox{\linewidth}{!}{%
\begin{tikzpicture}[
  >=latex, thick,
  block/.style={rectangle, draw, rounded corners, align=center, font=\small,
                minimum width=26mm, minimum height=8mm, inner sep=2pt, fill=blue!6},
  decision/.style={diamond, aspect=2, draw, align=center, inner sep=1pt,
                   minimum width=10mm, minimum height=10mm, font=\small, fill=orange!10},
  line/.style={-latex}
]

\node[block]    (input)    at (0,0)   {User Query $q$};
\node[block]    (delegate) at (3.5,0) {Belief-Guided\\Delegation\\(Thompson Sampling)};
\node[block]    (agent)    at (7,0)   {Selected Agent $A_{i^\star}$};
\node[block]    (judge)    at (10.8,0){LLM + Programmatic Judge};
\node[decision] (verdict)  at (14,0)  {Success?};

\node[block]    (memory)   at (14,2.2) {Update Posterior\\\& Store in Memory};

\node[block]    (refine)   at (14,-2.2){Refine Query \& Re-route};

\node[block]    (agg)      at (5.3,-2.2) {Aggregate Candidates\\(Selection + Evidence)};
\node[block]    (final)    at (5.3,2.2)  {Final Answer};

\draw[line] (input) -- (delegate);
\draw[line] (delegate) -- (agent);
\draw[line] (agent) -- (judge);
\draw[line] (judge) -- (verdict);

\draw[line] (verdict.north) -- (memory.south);
\draw[line] (memory.west) -- (final.east);

\draw[line] (verdict.south) -- (refine.north);
\draw[line] (refine.west) -- (agg.east);
\draw[line] (agg.north) -- (final.south);

\end{tikzpicture}%
}
\caption{\textbf{System architecture of \method.} Queries pass through belief-guided delegation, agent execution, and judge evaluation. Upon success (top path), the posterior is updated and stored in memory before producing the final answer. Upon failure (bottom path), the query is refined and re-routed, with candidates aggregated using evidence-based selection.}
\label{fig:flowchart}
\end{figure*}
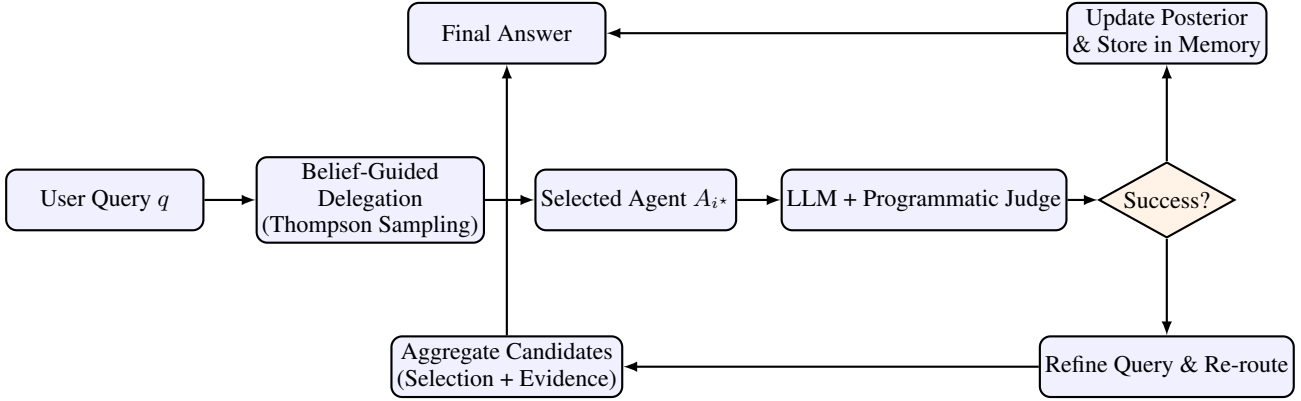

We formalize multi-agent coordination as a training-free probabilistic control loop. Given a task $T$ with query $q$ and a population of $N$ heterogeneous agents $A=\{A_1,\dots,A_N\}$, each agent $A_i$ possesses an unobserved task-conditional competence $\theta_i \in [0,1]$. The controller maintains a Beta posterior $\theta_i \sim \BetaD(\alpha_i,\beta_i)$, updated online from binary feedback $y\in\{0,1\}$ indicating success or failure in the evaluated results. Unless otherwise stated, priors are $\alpha_0=\beta_0=1$ (uninformative) or initialized by the memory-aware scheme in Section~\ref{sec:memory}. The overall decision-making pipeline of \method, including belief-guided delegation, recursive reflection, and memory-based adaptation, is illustrated in Figure~\ref{fig:flowchart}. This flowchart highlights how success cases are propagated through posterior updates and memory, whereas failures trigger refinement, aggregation, and rerouting until a satisfactory solution is produced.

Unlike standard bandit formulations, \method\ operates over compositional multi-agent reasoning trajectories where rewards are delayed, aggregated, and non-attributable to individual actions.

\paragraph{Interpretation of Agent Beliefs.}
The belief parameter $\theta_i$ does not represent the probability that agent $A_i$ can independently
solve the task. Instead, $\theta_i$ models the probability that invoking agent $A_i$ at the current
recursion step yields a \emph{net-positive marginal contribution} relative to the current candidate set.
This aligns Bernoulli feedback with compositional multi-agent tasks and avoids counterfactual blame
assignment to individual agents.

\subsection{Belief-Guided Delegation via Thompson Sampling (Core Policy)}
At recursion depth $d$ for query $q_d$, \method\ treats agent selection as a multi-armed bandit problem and applies Thompson sampling:
\begin{align}
\hat{\theta}_i &\sim \BetaD(\alpha_i,\beta_i),\qquad
i^\star \in \arg\max_i \hat{\theta}_i.
\end{align}
The selected agent $A_{i^\star}$ proposes a candidate solution. After judging, the posterior updates are in closed form:
\begin{align}
\alpha_{i^\star} \leftarrow \alpha_{i^\star} + y,\qquad
\beta_{i^\star} \leftarrow \beta_{i^\star} + (1-y),
\end{align}
where $y=\mathbb{I}[\text{success}]$. Thompson sampling provides a principled exploration--exploitation trade-off: agents with higher posterior means are preferred while uncertainty induces exploration.

We denote the posterior mean as $\mu_i = \frac{\alpha_i}{\alpha_i+\beta_i}$.

\subsection{Self-Reflection and Judging}
Each candidate is evaluated by a judge $J$ who delivers a calibrated binary verdict $E\in\{\text{\textsc{success}},\text{\textsc{failure}}\}$. Two evidence channels are integrated:
\begin{enumerate}[leftmargin=1.4em]
\item \textbf{Programmatic metrics}, when available (e.g., EM / F1, supporting facts F1, unit test pass rate), which short-circuit the success of unambiguous positives.
\item \textbf{LLM adjudication}, which produces a binary decision and a brief rationale when the metrics are absent or inconclusive.
\end{enumerate}
We calibrate $J$ on a small labeled set ($N\!\approx\!200$) to estimate FP/FN rates and set thresholds. The verdict is mapped to the Bernoulli feedback driving Bayesian updates:
\[
y \;=\; \mathbb{I}[E=\text{\textsc{success}}].
\]
This makes reflection the engine of both \emph{credit assignment} (posterior updates) and \emph{control} (whether to re-route).

\subsection{Text-Appropriate Aggregation}
\method\ aggregates by \emph{selection with evidence}, not averaging. We construct a set of candidates with rationales, score them using task metrics or retrieval-grounded entailment checks~\cite{Li25}, optionally run a bounded debate (up to two refinement turns) among top candidates, and select the best-supported answer.

For structured outputs (e.g., numeric fields, dates, JSON), we extract atomic fields, fuse them via competence-weighted voting using $\mu_i=\alpha_i/(\alpha_i+\beta_i)$ as trust weights, then regenerate a coherent output. This modality-aware design yields robustness across open-ended and structured tasks while maintaining interpretability.

\subsection{Recursive Re-Routing}
\label{sec:reroute}
If the judged outcome is \textsc{failure}, \method\ executes a budgeted recursive step: (i) update $\BetaD(\alpha_{i^\star},\beta_{i^\star})$ with $y=0$, (ii) refine the query with the judge's critique, and (iii) re-route via Thompson sampling to the next most promising expert. The recursion terminates upon any of the following:
\begin{enumerate}[leftmargin=1.4em]
\item \textbf{Success}: the judge returns \textsc{success};
\item \textbf{Depth limit}: reaching a maximum recursion depth $D$;
\item \textbf{Budget exhaustion}: cumulative cost (tokens, time) exceeding $B$;
\item \textbf{Plateau}: no judged improvement across recent iterations.
\end{enumerate}
This procedure is not full tree search (no rollout value estimation) but a lightweight, online mechanism that reliably recovers from local errors and discovers productive multi-agent chains with low overhead. The overall recursive procedure of \method, combining Bayesian delegation, self-reflection, query refinement, and memory-aware updating, is summarized in Algorithm~\ref{alg:method}.

\begin{algorithm}[t]
\caption{\method: Recursive Delegation and Reflection}
\label{alg:method}
\begin{algorithmic}[1]
\Require Query $q$, agents $A_1..A_N$ with $(\alpha_i,\beta_i)$, memory $\mathcal{M}$, judge $J$, max depth $D$, budget $B$, cooldown $r$
\State $x \gets \mathrm{embed}(q)$
\State initialize $(\alpha_i,\beta_i)$ via memory-aware priors from $\mathcal{M}$
\State $\mathrm{cool}[i]\gets 0$ for all $i$;\quad $\mathrm{spent}\gets 0$
\State $\mathcal{C}\gets \emptyset$ \Comment{set of all candidates with evidence}
\State $(\mathrm{best}, \mathrm{bestScore}) \gets (\varnothing, -\infty)$
\For{$d=1$ to $D$}
  \If{$\mathrm{spent} \ge B$} \textbf{break} \EndIf
  \State \textbf{Sampling:} for each $i$, draw $\hat{\theta}_i \sim \BetaD(\alpha_i,\beta_i)$ if $\mathrm{cool}[i]=0$ else set $\hat{\theta}_i\gets -\infty$
  \If{$\max_i \hat{\theta}_i=-\infty$} \Comment{all cooling; force one step of exploration}
      \State set $\mathrm{cool}[j]\gets 0$ for $j=\arg\max_i \mathrm{cool}[i]$ \Comment{or the smallest cooldown}
      \State \textbf{continue}
  \EndIf
  \State $i^\star \gets \arg\max_i \hat{\theta}_i$ \Comment{tie-break by larger $\alpha_i/(\alpha_i+\beta_i)$}
  \State $(\mathrm{cand}, \mathrm{usage}) \gets A_{i^\star}(q)$
  \State $\mathrm{spent} \mathrel{+}= \mathrm{usage}$
  \If{$\mathrm{spent} > B$} \textbf{break} \EndIf
  \State $\mathrm{score}_{\mathrm{prog}} \gets \mathrm{programmatic\_metrics}(q,\mathrm{cand})$
  \If{$\mathrm{score}_{\mathrm{prog}}$ is unambiguous positive}
      \State $E \gets \text{\textsc{success}}$;\quad $\rho \gets \text{``programmatic pass''}$
  \Else
      \State $(E,\rho,\mathrm{score}_{\mathrm{judge}}) \gets J(q,\mathrm{cand},\mathrm{score}_{\mathrm{prog}})$
  \EndIf
  \State $y \gets \mathbb{I}[E=\text{\textsc{success}}]$
  \State $\alpha_{i^\star}\mathrel{+}=y$;\quad $\beta_{i^\star}\mathrel{+}=(1-y)$
  \State $\mathcal{M}\gets \mathcal{M}\cup\{(x,i^\star,y,\rho,\mathrm{now}())\}$
  \State $\mathcal{C}\gets \mathcal{C}\cup\{(\mathrm{cand}, \mathrm{score}_{\mathrm{prog}}, \mathrm{score}_{\mathrm{judge}}, i^\star)\}$
  \State $\mathrm{candScore}\gets \mathrm{combine\_scores}(\mathrm{score}_{\mathrm{prog}}, \mathrm{score}_{\mathrm{judge}})$
  \If{$\mathrm{candScore} > \mathrm{bestScore}$}
      \State $(\mathrm{best}, \mathrm{bestScore}) \gets (\mathrm{cand}, \mathrm{candScore})$
  \EndIf
  \If{$y=1$}
      \State \Return $\mathrm{aggregate\_select}(\mathcal{C}, \{\mu_i=\alpha_i/(\alpha_i+\beta_i)\})$ \Comment{selection with evidence}
  \EndIf
  \State $\mathrm{cool}[i^\star] \gets r$;\quad \textbf{for all $j$:} $\mathrm{cool}[j]\gets \max(0, \mathrm{cool}[j]-1)$
  \State $q \gets \mathrm{refine}(q,\mathrm{cand},\rho)$
  \If{$\mathrm{plateau}(\mathcal{C}, k)$} \textbf{break} \Comment{no score improvement over last $k$ attempts}
  \EndIf
\EndFor
\If{$\mathcal{C}\neq\emptyset$}
    \State \Return $\mathrm{aggregate\_select}(\mathcal{C}, \{\mu_i\})$ \Comment{best-so-far with evidence checks}
\Else
    \State \Return \textsc{failure} \Comment{no candidate produced}
\EndIf
\end{algorithmic}
\end{algorithm}

\subsection{Design Rationale}
Several design choices in Algorithm~\ref{alg:method} merit justification.

\textbf{Why Thompson sampling over UCB or $\varepsilon$-greedy?}
Thompson sampling naturally scales exploration to posterior uncertainty: when beliefs are diffuse (early rounds), it explores broadly; as posteriors concentrate, it exploits. UCB requires explicit tuning of the confidence parameter, and $\varepsilon$-greedy explores uniformly regardless of evidence, wasting budget on agents already known to be weak.

\textbf{Why selection over averaging?}
Averaging multiple agent outputs degrades quality when agents have heterogeneous competence---a strong answer is diluted by weak ones. Selection with evidence preserves the best candidate while the bounded debate (two rounds) allows targeted refinement without the noise of full ensembling.

\textbf{Why cooldown?}
Without cooldown, a single false-positive verdict can cause the system to repeatedly select the same agent. The cooldown $r$ forces exploration of alternatives after each selection, complementing the stochastic exploration of Thompson sampling with a deterministic diversity mechanism.

\textbf{Why binary feedback?}
Richer feedback (e.g., scalar scores) could accelerate learning but requires a well-calibrated scoring model. Binary success/failure is robust to judge miscalibration---the Beta-Bernoulli posterior is a conjugate update that remains valid even under moderate label noise (Theorem~\ref{thm:noisy-regret}).

\subsection{Memory-Aware Priors and Cold-Start Mitigation}
\label{sec:memory}
To reduce early-round inefficiency, we seed priors with similarity- and recency-weighted historical outcomes.

\begin{align}
\alpha_i &\leftarrow \alpha_0 + \sum_{m} K(x_d,x_m)\, y_m \, w_{\Delta t_m}, \\
\beta_i  &\leftarrow \beta_0  + \sum_{m} K(x_d,x_m)\, (1-y_m)\, w_{\Delta t_m},
\end{align}
where $x_d=\mathrm{embed}(q_d)$, $K$ is a task-similarity kernel (e.g., cosine over sentence embeddings), and $w_{\Delta t_m}=\exp(-\lambda \Delta t_m)$ applies temporal decay.   This initialization biases competence posteriors toward agents that recently succeeded on similar tasks while preserving adaptability under drift.

\paragraph{Remark (Interpretability).}
The beta parameters $(\alpha_i,\beta_i)$, the rationales of the judges $\rho$, and the selection history (agents chosen, verdicts, costs) constitute an auditable decision trail, which facilitates error analysis and responsible deployment.

\subsection{Robustness Considerations}
The controller relies on binary feedback from a judge, which may be imperfect.
To mitigate this, \textsc{REDEREF} incorporates several robustness properties:
(i)~the Beta posterior is inherently resilient to occasional mislabeled feedback because individual observations have diminishing influence as evidence accumulates;
(ii)~the cooldown mechanism prevents the system from repeatedly exploiting a single agent based on a single lucky positive verdict;
and (iii)~ensemble judges (majority voting over three independent judges) can further reduce sensitivity to individual judge errors at modest additional cost.

\subsection{Regret Under Noisy Feedback}
We now quantify the cost of imperfect judging.
Let $\varepsilon_{\mathrm{FP}}=\Pr[E{=}\text{\textsc{success}}\mid \text{true fail}]$ and
$\varepsilon_{\mathrm{FN}}=\Pr[E{=}\text{\textsc{failure}}\mid \text{true success}]$ denote the judge's
false-positive and false-negative rates, and define the \emph{discrimination margin}
$\delta=1-\varepsilon_{\mathrm{FP}}-\varepsilon_{\mathrm{FN}}>0$.

\begin{theorem}[Regret under noisy judge feedback]\label{thm:noisy-regret}
Consider $N$ agents with true competences $\theta_1,\dots,\theta_N$ and a judge
with class-conditional error rates $(\varepsilon_{\mathrm{FP}},\varepsilon_{\mathrm{FN}})$
satisfying $\delta=1-\varepsilon_{\mathrm{FP}}-\varepsilon_{\mathrm{FN}}>0$.
Then the Bayesian regret of Thompson sampling over $T$ rounds satisfies
\[
  R(T)
  \;=\;
  \frac{1}{\delta}\,\widetilde{R}(T)
  \;=\;
  O\!\Bigl(\frac{\sqrt{NT\log T}}{\delta}\Bigr),
\]
where $\widetilde{R}(T)$ is the regret of Thompson sampling on an equivalent
noiseless problem with transformed competences
$\tilde{\theta}_i=\delta\,\theta_i+\varepsilon_{\mathrm{FP}}$.
\end{theorem}

\begin{corollary}
As $\delta\to 0$ (uninformative judge), regret diverges; as
$\delta\to 1$ (perfect judge), we recover the standard
$O(\sqrt{NT\log T})$ bound of~\citet{AgrawalGoyal13}.
\end{corollary}

\noindent The proof is given in Appendix~\ref{app:noisy-regret-proof}. Theorem~\ref{thm:noisy-regret} shows that judge noise inflates regret by exactly $1/\delta$, providing a precise budget for investing in judge quality versus accepting routing suboptimality.

\subsection{Summary and Preview}
The framework provides one theoretical guarantee and three empirically testable properties. Theorem~\ref{thm:noisy-regret} bounds regret under noisy feedback as $O(\sqrt{NT\log T}/\delta)$, establishing that routing quality degrades gracefully with judge noise. The remaining properties---that belief-guided routing improves efficiency over uninformed routing (H1), that posteriors concentrate on domain experts (H2), and that the routing policy adapts to agent degradation (H3)---are validated experimentally in Sections~\ref{sec:exp}--\ref{sec:results}.

\section{Experimental Design}
\label{sec:exp}

Because recursive retry already achieves high success on split-knowledge tasks, we treat task success as a \emph{constraint} rather than the primary optimization objective. Our evaluation is therefore a controlled ablation: we isolate the effect of Bayesian belief-guided routing within an otherwise identical recursive delegation loop, measuring whether informed agent selection improves token usage, agent calls, and time-to-first-success relative to uninformed (uniform random) selection.

\begin{figure*}[t]
\centering
\resizebox{\linewidth}{!}{%
\begin{tikzpicture}[
  >=latex, thick,
  box/.style={rectangle, draw, rounded corners, align=center, font=\small,
              minimum width=28mm, minimum height=8mm, fill=blue!6},
  metric/.style={rectangle, draw, rounded corners, align=left, font=\scriptsize,
                 minimum width=36mm, minimum height=6mm, fill=green!5},
  line/.style={-latex}
]

\node[box] (rq1) at (0,0) {RQ1: Composition Efficiency};
\node[box] (rq2) at (0,-2.2) {RQ2: Specialization};
\node[box] (rq3) at (0,-4.4) {RQ3: Adaptability};

\node[box, right=3.8cm of rq1] (h1) {H1: Belief-guided routing reduces\\ cost at matched success};
\node[box, right=3.8cm of rq2] (h2) {H2: Belief scores diverge,\\ reflecting specialization};
\node[box, right=3.8cm of rq3] (h3) {H3: Routing adapts to impaired\\ agents in real time};

\node[metric, right=5.2cm of h1] (m1) {
\begin{tabular}{@{}l@{}}
-- Success Rate (constraint) \\
-- Tokens \\
-- Agent Calls \\
-- Time-to-first-success
\end{tabular}};
\node[metric, right=5.2cm of h2] (m2) {
\begin{tabular}{@{}l@{}}
-- Belief Score Evolution \\
-- Rounds to Expert Selection
\end{tabular}};
\node[metric, right=5.2cm of h3] (m3) {
\begin{tabular}{@{}l@{}}
-- Impairment Test (pre/post) \\
-- Belief Score Decline \\
-- Contribution Frequency \\
-- Final Output Score
\end{tabular}};

\draw[line] (rq1) -- (h1);
\draw[line] (rq2) -- (h2);
\draw[line] (rq3) -- (h3);

\draw[line] (h1) -- (m1);
\draw[line] (h2) -- (m2);
\draw[line] (h3) -- (m3);

\end{tikzpicture}%
}
\caption{\textbf{Evaluation framework.} Research questions, testable hypotheses, and associated metrics. Each dimension targets a distinct property of the routing mechanism.}
\label{fig:rq-flow}
\end{figure*}
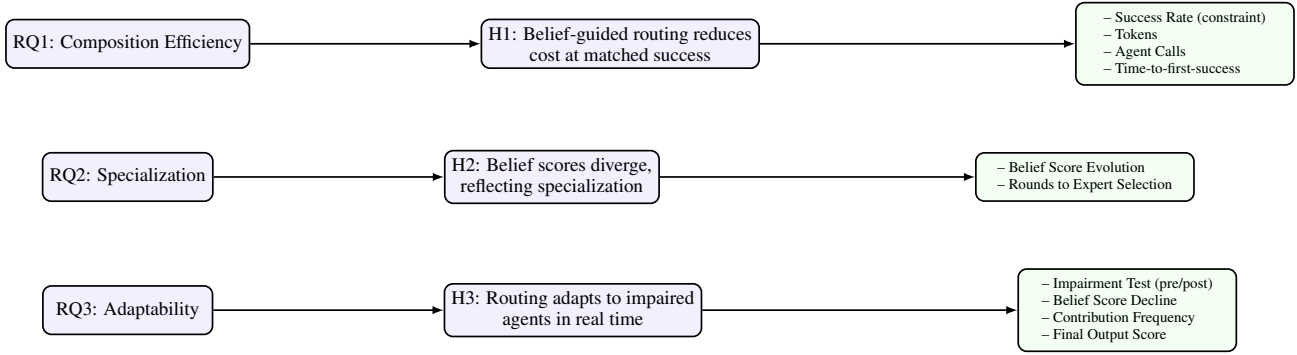

\subsection{Research Questions}
We organize evaluation around three dimensions (Figure~\ref{fig:rq-flow}):
\begin{itemize}[leftmargin=1.4em]
    \item[\textbf{RQ1}] \textbf{Composition efficiency (H1):} At matched task success, does belief-guided routing reduce token usage, agent calls, and time-to-first-success compared to random recursive delegation?
    \item[\textbf{RQ2}] \textbf{Specialization (H2):} Do belief scores diverge over task sequences, with domain experts being preferentially selected for related tasks?
    \item[\textbf{RQ3}] \textbf{Adaptability (H3):} When a previously reliable agent is impaired mid-experiment, does the routing policy detect the degradation and redirect queries?
\end{itemize}

\subsection{Tasks and Agent Population}
We constructed a benchmark of split-knowledge tasks via a three-stage pipeline (full details in Appendix): (i)~domain-specific question generation for each agent (20--22 questions each), (ii)~multi-agent task synthesis by sampling 4--6 agents and composing questions that require non-redundant contributions from each, and (iii)~iterative sampling until the question pool was exhausted. By construction, each task requires contributions from multiple agents and cannot be solved by any single agent alone.

The agent population comprises 6 specialist RAG agents (biology, finance, law, medicine, electrical engineering, mathematics) with retrieval over curated knowledge bases, and 10 generalist conversational agents (fitness, literature, technology, geography, storytelling, politics, academics, career guidance, trivia, daily planning). All agents use Azure OpenAI endpoints with fixed random seeds for reproducibility. Agents are initialized with uniform priors $\alpha_0=\beta_0=1$ unless memory-aware initialization is enabled.

\subsection{Baseline}
Since split-knowledge tasks are unsolvable by any single agent, single-agent baselines are excluded by task design. The comparison instead isolates the routing mechanism:
\begin{itemize}[leftmargin=1.4em]
    \item \textbf{Random Delegation:} The full recursive loop (judging, re-routing, aggregation, budget limits) is retained identically, but agent selection at each step is uniformly random. This ensures that any observed difference is attributable to belief-guided routing rather than to recursion, judging, or aggregation.
\end{itemize}

\subsection{Evaluation Metrics}
\textbf{Performance (constraint):} Task success rate (final score $\geq 85$ as judged by a calibrated LLM) and output quality, reported to confirm that both methods achieve comparable success.
\textbf{Efficiency (H1):} Token usage, agent calls, and time-to-first-success, reported as ratios normalized to Random Delegation (lower is better).
\textbf{Specialization (H2):} Belief score trajectories over task sequences and rounds-to-expert-selection for domain-specific tasks.
\textbf{Adaptability (H3):} In an impairment test (50 normal followed by 50 degraded tasks for one agent), we track belief score decline, contribution frequency shift, and system output quality.
Full metric definitions are provided in the Appendix.

\section{Results and Analysis}
\label{sec:results}

We summarize both \emph{effectiveness} (success, quality) and \emph{efficiency/robustness} (tokens, calls, latency, routing efficiency, and adaptation). Although recursion alone is surprisingly strong on our split-knowledge suite, \method\ consistently shifts the \emph{efficiency frontier}: at matched success it uses fewer tokens and agent calls, reaches the first success faster, and adapts under drift, whereas random delegation wastes capacity on unreliable agents.

\subsection{Overall Performance and Efficiency}
Because recursive retry already saturates success on our split-knowledge tasks, we report success
alongside efficiency metrics (tokens, agent calls, and time-to-first-success). Random recursive delegation
matches \method\ on success rate, but \method\ substantially reduces wasted calls by prioritizing agents
with higher historical marginal contribution. We report tokens, agent calls, and time-to-success normalized to Random Delegation (1.00$\times$).

\begin{figure*}[t]
\centering
\begin{subfigure}[t]{0.48\linewidth}
  \centering
  \includegraphics[width=\linewidth]{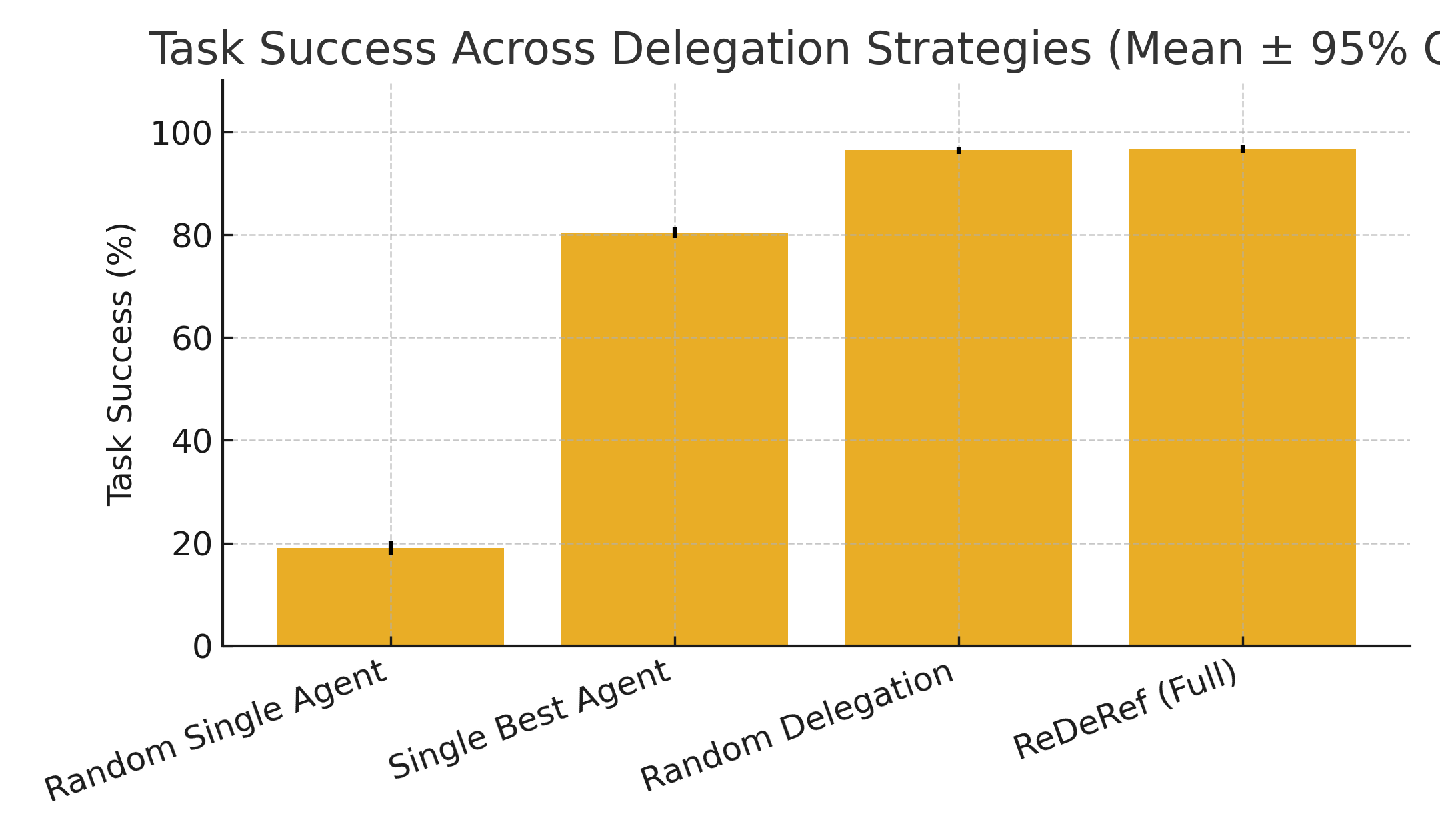}
  \caption{Task success across delegation strategies (mean $\pm$ 95\% CI).}
  \label{fig:success-ci}
\end{subfigure}\hfill
\begin{subfigure}[t]{0.48\linewidth}
  \centering
  \includegraphics[width=\linewidth]{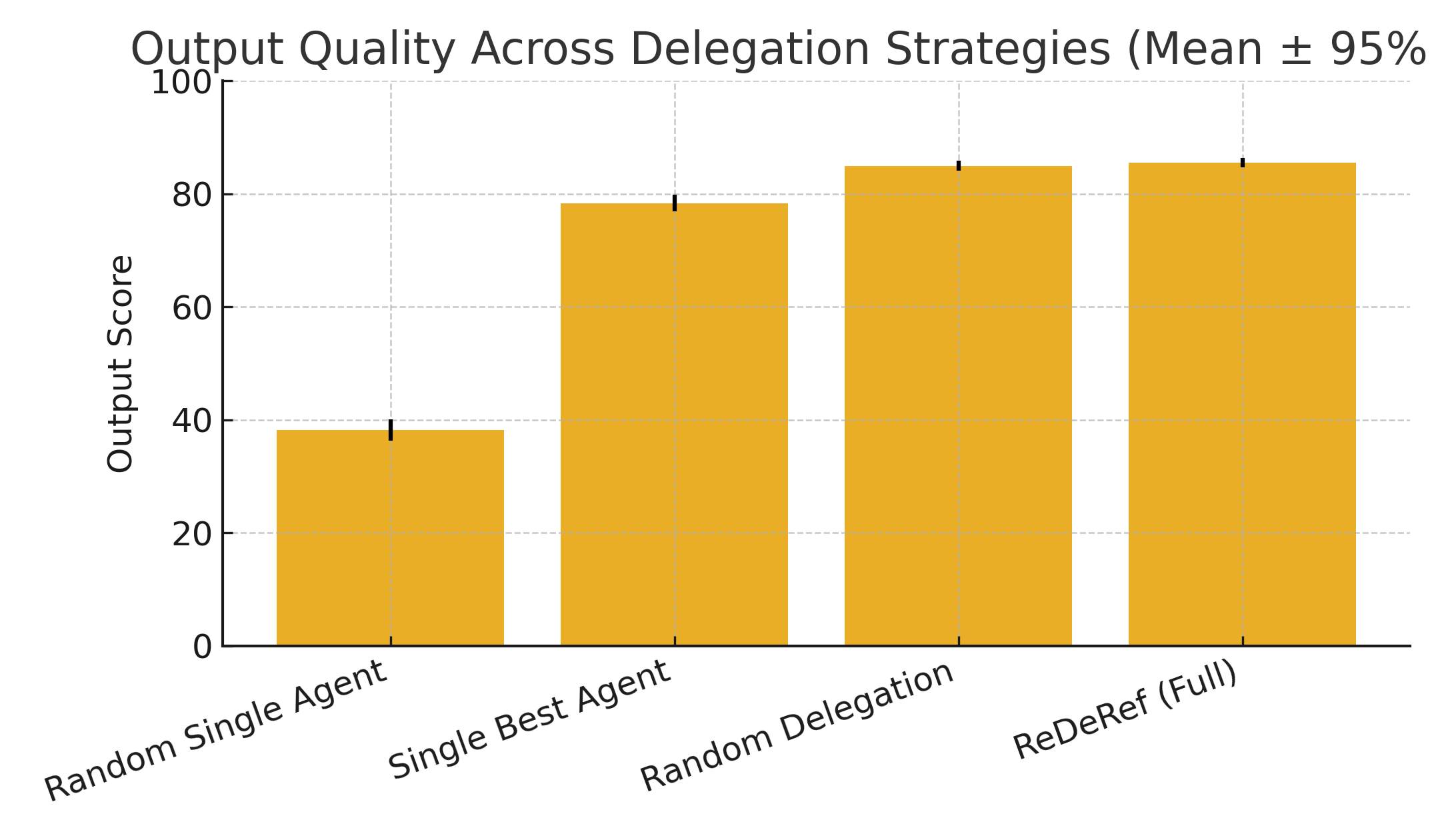}
  \caption{Output quality across delegation strategies (mean $\pm$ 95\% CI).}
  \label{fig:quality-ci}
\end{subfigure}
\caption{\textbf{Performance comparison across delegation strategies.}
(a) Task success rates (mean $\pm$ 95\% CI) are saturated by recursive retry.
(b) Output quality at convergence. \method\ primarily improves \emph{efficiency} (fewer calls/tokens)
while maintaining comparable success. Error bars represent 95\% confidence intervals.}
\label{fig:ci-combined}
\end{figure*}

Figure~\ref{fig:ci-combined} complements the table by visualizing mean performance with confidence intervals, reinforcing that recursive retry saturates success, while \method\ improves efficiency by reducing wasted calls and tokens.

\subsection{Composition Dynamics and Efficiency}
\begin{figure*}[t]
\centering
\begin{subfigure}[t]{0.58\linewidth}
  \centering
  \includegraphics[width=\linewidth]{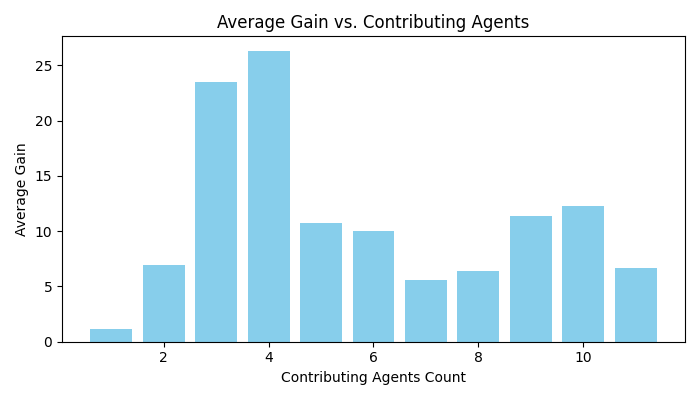}
  \caption{Average quality improvement vs.\ number of contributing agents. Collaboration among 3--5 strong contributors yields the largest quality improvements.}
  \label{fig:gain-combined1}
\end{subfigure}\hfill
\begin{subfigure}[t]{0.38\linewidth}
  \centering
  \includegraphics[width=\linewidth]{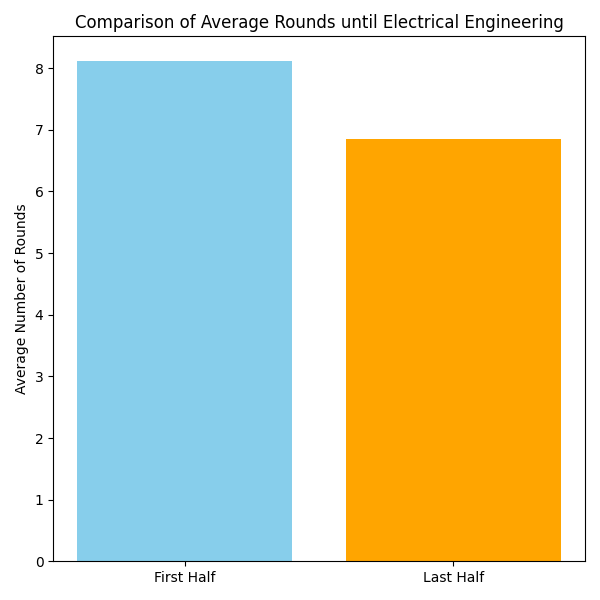}
  \caption{Average rounds until Electrical Engineering expert was selected across task halves.}
  \label{fig:gain-combined2}
\end{subfigure}
\caption{\textbf{Collaboration dynamics and specialization.} (a) Average quality gain versus number of contributing agents, showing peak composition efficiency at 3--5 agents. (b) Decline in rounds required to select the Electrical Engineering expert across task sequence, demonstrating specialization via posterior concentration.}
\label{fig:composition-dynamics}
\end{figure*}

Beyond aggregate performance, we examined how collaboration among multiple agents contributes to final outcomes. As shown in Figure~\ref{fig:gain-combined1}, average score improvements are highest when 3--5 agents contribute meaningfully to the final output. This validates the intuition that composition efficiency---not just redundancy---drives quality: a small coalition of diverse experts yields larger improvements than either a single agent or overly diffuse collaboration.

\subsection{Ablation: Role of Belief Updates}
To isolate the impact of belief-driven delegation, we compared the full system to a variant in which agents were selected uniformly at random during recursive rerouting. As shown in Table~\ref{table:routing_behavior}, disabling belief updates leads to a 17.2\% increase in the use of underperforming agents and an 8.2\% increase in exploratory attempts before reaching a successful outcome. Both differences are statistically significant under paired bootstrap tests ($p < 0.01$). While the number of useful contributors remains similar, the random policy wastes capacity on unreliable agents, underscoring that experience-driven learning enhances both efficiency and robustness.

\begin{table}[t]
\centering
\caption{Efficiency comparison at matched task success. Tokens, Agent Calls, and Time-to-Success are reported as ratios normalized to Random Delegation; lower is better.}
\resizebox{\columnwidth}{!}{%
\begin{tabular}{@{}lcccc@{}}
\toprule
\textbf{Method} & \textbf{Success (\%)} & \textbf{Tokens ($\downarrow$)} & \textbf{Agent Calls ($\downarrow$)} & \textbf{Time-to-Success ($\downarrow$)} \\
\midrule
Random Delegation & 96.46 & 1.00$\times$ & 1.00$\times$ & 1.00$\times$ \\
\method\ (Full) & \textbf{96.65} & \textbf{0.72$\times$} & \textbf{0.83$\times$} & \textbf{0.81$\times$} \\
\bottomrule
\end{tabular}}
\label{table:results}
\end{table}

\begin{table}[t]
\centering
\caption{Routing behavior with and without belief updates (mean $\pm$ 95\% CI).}
\resizebox{\columnwidth}{!}{%
\begin{tabular}{@{}lccc@{}}
\toprule
\textbf{Method} & \textbf{Bad Agent Uses} & \textbf{Good Agent Uses} & \textbf{Total Attempted} \\
\midrule
Random Delegation & 6.62 $\pm$ 0.4 & 4.92 $\pm$ 0.2 & 11.54 $\pm$ 0.5 \\
\method\ (Full) & \textbf{5.65 $\pm$ 0.3} & 5.01 $\pm$ 0.2 & \textbf{10.67 $\pm$ 0.4} \\
\bottomrule
\end{tabular}}
\label{table:routing_behavior}
\end{table}

\subsection{Specialization}
We next examined whether agents developed stable specializations over time. In a sequence of 55 electrical-engineering tasks, the median belief score for the domain expert rose from 0.50 ($\pm$0.02) to 0.84 ($\pm$0.03). Concurrently, the average number of rounds required before this expert was selected declined from 8.11 ($\pm$0.4) to 6.86 ($\pm$0.3), as shown in Figure~\ref{fig:gain-combined2}. This trajectory demonstrates that \method{} not only improves aggregate performance but also learns to preferentially route domain-specific queries to the most competent agents---consistent with specialization via posterior concentration (supporting H2).

\subsection{Adaptability to Agent Impairment}
\begin{figure}[t]
    \centering
    \includegraphics[width=\columnwidth]{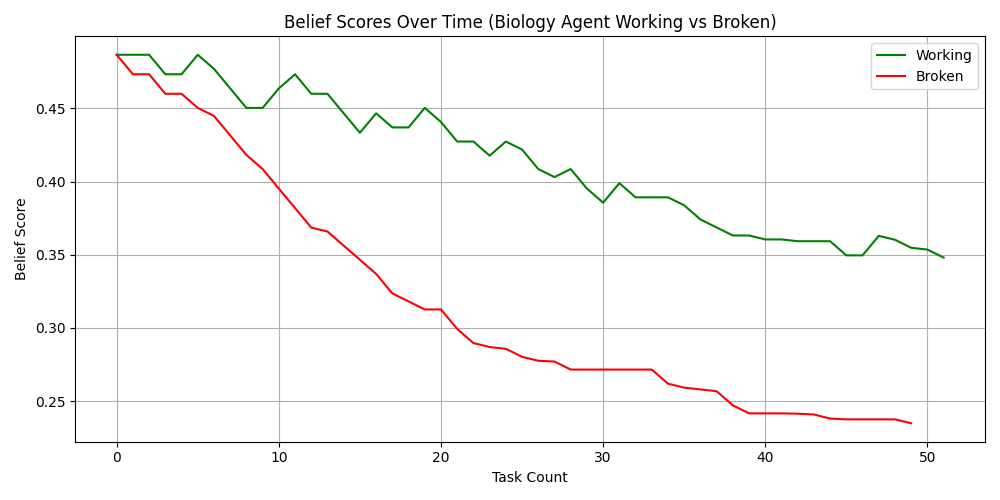}
    \caption{\textbf{Adaptability under agent impairment.} Belief score trajectories for the Biology agent under normal versus systematically impaired conditions. The system rapidly detects degradation and down-weights the compromised agent, demonstrating real-time adaptability.}
    \label{fig:overlay_belief_scores}
\end{figure}

To evaluate adaptability (H3), we conducted an impairment test in which the \texttt{Biology\_RAG\_Agent} was replaced with systematically poor outputs after the first 50 tasks. As shown in Table~\ref{tab:agent_impairment_test}, the agent's belief score decreased by almost 50\%, and its contributions were eliminated completely in subsequent tasks. While the overall output quality decreased modestly (84.52 $\pm$ 0.7 to 81.80 $\pm$ 0.9), the system dynamically reallocated queries to other competent agents, preventing catastrophic degradation. This rapid adjustment is also illustrated in Figure~\ref{fig:overlay_belief_scores}, which shows diverging belief trajectories for the impaired and healthy agents.

\begin{table}[t]
\centering
\caption{Adaptability under impairment: Biology agent performance before and after enforced degradation (mean $\pm$ 95\% CI).}
\begin{tabular}{@{}lcc@{}}
\toprule
\textbf{Metric} & \textbf{Normal} & \textbf{Impaired} \\
\midrule
Average Belief Score & 0.35 $\pm$ 0.02 & 0.23 $\pm$ 0.03 \\
Agent Contributions & 12 $\pm$ 1.1 & 0 \\
Final Output Score & 84.52 $\pm$ 0.7 & 81.80 $\pm$ 0.9 \\
\bottomrule
\end{tabular}
\label{tab:agent_impairment_test}
\end{table}

\subsection{Qualitative Dynamics}
Qualitative inspection further illustrates \method's recursive dynamics. In one representative task, the system initially delegated to an electrical-engineering agent, which produced a technically correct but incomplete design. After a failure judgment, \method{} re-delegated to a narrative-oriented agent, which supplied the missing community-education perspective. The final aggregated response integrated both technical and social dimensions and was judged successful. Such trajectories exemplify how recursive re-routing enables the system to synthesize complementary expertise and recover from early missteps.

\paragraph{Summary.}
In summary, belief-guided routing reduces token usage, agent calls, and time-to-success at matched success rates (H1), belief scores concentrate on domain experts over task sequences (H2), and the routing policy adapts when an agent is impaired mid-experiment (H3). All three hypotheses from Section~\ref{sec:exp} are supported.

\section{Discussion and Conclusion}
We introduced \method, a lightweight, training-free controller for multi-agent LLM systems that addresses three persistent challenges in multi-agent collaboration: dynamic task routing, long-horizon credit assignment, and cold-start inefficiency. By combining Thompson-sampling delegation, reflection-driven updates, and memory-aware priors, \method{} maintains saturated task success while improving routing efficiency and robustness with minimal overhead.

\paragraph{Key insights.} Our findings support a growing body of evidence that lightweight, interpretable mechanisms can rival or surpass resource-intensive pipelines. The probabilistic control formulation resonates with models of agent trust~\cite{Wu21} and bandit methods for online decision-making~\cite{Chapelle11}, while avoiding the instability and sample inefficiency of deep RL~\cite{Mnih15}. The recursive loop mirrors ideas from search-based reasoning such as Monte Carlo Tree Search~\cite{Browne12}, yet operates at a fraction of the computational cost. In ablations, removing belief-guided routing increased poor-agent usage by 17.2\% and wasted attempts, underscoring the critical importance of online feedback for efficient coordination.

\paragraph{Limitations.} \method{} depends critically on judge reliability; biases in machine adjudication are well documented~\cite{Karpinska21} and can distort competency updates, potentially entrenching suboptimal routing decisions. Sequential recursion can add latency in long tasks, limiting applicability in real-time scenarios. Early cold-start behavior resembles random delegation before sufficient evidence accumulates to differentiate agent competencies. These limitations highlight the need for robust judge calibration, potential parallelization of candidate generation, and more informative prior initialization strategies.

\paragraph{Broader implications.} The transparency of \method{} is advantageous for responsible AI deployment: Beta posteriors, judge verdicts, and natural language rationales create an auditable decision trail~\cite{Doshi17}, contrasting sharply with opaque RL policies or heavily fine-tuned black-box controllers. However, adaptive down-weighting based on early evidence could prematurely exclude competent agents, particularly in scenarios with temporary performance fluctuations. Periodic recalibration and ensemble judge mechanisms are therefore important for reliable deployment in sensitive domains.

\paragraph{Future directions.} Several extensions warrant investigation. Hybridizing Bayesian delegation with model-based reinforcement learning could combine fast online adaptation with long-term strategic planning~\cite{Silver17}. Expanding beyond binary success/failure feedback into richer error taxonomies (e.g., reasoning errors, factual inaccuracies, incomplete coverage) would enable more fine-grained credit assignment and targeted agent improvement. Parallel candidate generation could significantly reduce latency while maintaining bounded computational budgets. Finally, programmatic verifiers and retrieval-grounded evaluation models~\cite{min2023factscore} offer promising directions for improving judge reliability and reducing calibration overhead.

\paragraph{Conclusion.} Robust collaboration in LLM collectives does not require complex black-box architectures or extensive training. Simple probabilistic mechanisms---belief-guided delegation, calibrated reflection,
and memory-aware priors---can transform collections of independent agents into cohesive,
adaptive systems. This ``fast and frugal'' approach offers a scalable, interpretable, and practically deployable path forward for multi-agent LLM research and real-world applications.

\bibliography{references}
\bibliographystyle{icml2026}

\appendix
\section{Appendix}
\subsection{Evaluation Task Generation}
To ensure diversity and rigor, we constructed a multi-stage pipeline to generate tasks requiring multi-agent collaboration.

\subsubsection{Stage 1: Single-Agent Question Generation}
We first generated domain-specific questions tailored to two classes of agents:
\begin{itemize}
    \item \textbf{RAG Agents}: Retrieval-augmented models grounded in curated domain datasets (e.g., biology, finance, medicine)~\cite{huggingface}.
    \item \textbf{Conversational Agents}: Prompt-based agents covering domains such as career guidance, fitness, or literature.
\end{itemize}
Each agent was tasked with producing 20--22 realistic questions, paired with a \textit{Model Context Protocol} (MCP) capturing intent, tools, and plausible follow-ups. All outputs were archived in structured JSON format.

\subsubsection{Stage 2: Multi-Agent Task Synthesis}
Single-agent questions were combined into multi-agent tasks by sampling 4--6 diverse agents and synthesizing 15 composite tasks per batch. Each task was required to:
\begin{itemize}
    \item Necessitate distinct, non-redundant contributions from each agent,
    \item Require multi-layered reasoning (planning, analysis, execution),
    \item Include a merged MCP integrating the intents, tools, and follow-ups of all contributing agents.
\end{itemize}
Examples include generating healthcare policy reports using medical, political, and geographic agents, or constructing recovery plans with biology, fitness, finance, and scheduling agents.

\subsubsection{Stage 3: Iterative Sampling}
This process was repeated until the pool of questions was exhausted, yielding a benchmark suite explicitly designed to enforce distributed reasoning. Each task was unsolvable by any single agent, thereby ensuring collaborative evaluation.

\subsection{Extended Metric Definitions}
This section provides full definitions of the evaluation metrics summarized in Section~\ref{sec:exp}.

\subsubsection{Performance}
\begin{itemize}
    \item \textbf{Task Success Rate:} Percentage of tasks achieving a final answer score $\geq 85$ as judged by an LLM.
    \item \textbf{Agents Attempted:} Average number of agents invoked before reaching a solution.
\end{itemize}

\subsubsection{Specialization (H2)}
\begin{itemize}
    \item \textbf{Belief Score Evolution:} Temporal trajectories of $b_i$ for domain experts (e.g., Math agent).
    \item \textbf{Rounds to Expert Selection:} Average delegation steps before the correct expert is chosen for domain-specific tasks, expected to decline as belief updates accumulate.
\end{itemize}

\subsubsection{Adaptability (H3)}
\begin{itemize}
    \item \textbf{Agent Impairment Test:} Across 100 tasks, one agent (e.g., Biology) operates normally for the first 50, then is forced to fail for the next 50.
    \item \textbf{Belief Score Decline:} Comparison of belief trajectories pre- and post-impairment.
    \item \textbf{Contribution Frequency:} Task participation counts across both phases.
    \item \textbf{System Output Score:} Average final answer score across both segments, reflecting resilience to impaired agents.
\end{itemize}

\subsubsection{Composition Efficiency (H1)}
\begin{itemize}
    \item \textbf{Time-to-first-success:} Number of delegation rounds required to produce the first judged-successful candidate.
    \item \textbf{Marginal contribution efficiency:} Improvement in output score per token consumed by each contributing agent.
\end{itemize}

 \subsection{Implementation Details}

  \subsubsection{Framework Architecture}
  The \method\ framework is implemented as a modular pipeline consisting of six core components: \texttt{WorkflowManager}, \texttt{BayesianDelegator}, \texttt{SelfReflectionStep}, \texttt{InfoMergeStep}, \texttt{MemoryUpdateStep}, and \texttt{RecursiveRouting}. The
  \texttt{WorkflowManager} serves as the main orchestrator, coordinating agent delegation, reflection, memory updates, and recursive re-routing according to Algorithm~\ref{alg:method}.

  \paragraph{Core Classes and Data Structures.}
  The central \texttt{Task} class maintains task state including query text, agent outputs, reflection scores, memory updates, and completion status. Agent performance is tracked through belief parameters $(\alpha_i, \beta_i)$, cooldown counters, and historical success rates. The
  framework supports both Bayesian delegation via Thompson sampling and traditional LLM-based ranking through a configurable \texttt{use\_bayesian} parameter.

  \subsubsection{Bayesian Delegation Implementation}
  Thompson sampling is implemented in the \texttt{BayesianDelegator} class with the following key features:
  \begin{itemize}[leftmargin=1.4em]
      \item \textbf{Agent Selection}: For each agent $i$, sample $\hat{\theta}_i \sim \mathrm{Beta}(\alpha_i, \beta_i)$ and select $i^* = \arg\max_i \hat{\theta}_i$ (subject to cooldown constraints).
      \item \textbf{Binary Updates}: After judging, update $\alpha_{i^*} \leftarrow \alpha_{i^*} + y$ and $\beta_{i^*} \leftarrow \beta_{i^*} + (1-y)$ where $y \in \{0,1\}$ indicates success/failure.
      \item \textbf{Cooldown Mechanism}: Agents are temporarily excluded for $r$ rounds after selection to encourage exploration. If all agents are cooling, the framework forces exploration by selecting the agent with the smallest remaining cooldown.
      \item \textbf{Belief Persistence}: Agent beliefs are stored in JSON format and loaded across sessions to maintain long-term memory.
  \end{itemize}

  \subsubsection{Memory-Aware Prior Initialization}
  Historical performance data is used to initialize belief priors via similarity-weighted aggregation:
  Let $s_m = K(\mathrm{embed}(q),\, \mathrm{embed}(q_m)) \cdot \exp(-\lambda \Delta t_m)$. Then:
  \begin{align}
  \alpha_i &\leftarrow \alpha_0 + \textstyle\sum_{m \in \mathcal{M}} s_m \cdot y_m, \\
  \beta_i  &\leftarrow \beta_0  + \textstyle\sum_{m \in \mathcal{M}} s_m \cdot (1 - y_m),
  \end{align}
  where $K(\cdot, \cdot)$ is cosine similarity over sentence embeddings, $\Delta t_m$ is task recency, and $\lambda = 0.1$ controls temporal decay. This initialization reduces cold-start inefficiency by biasing selection toward agents with historically strong performance on similar
  tasks.

  \subsubsection{Multi-Layered Judge System}
  The \texttt{SelfReflectionStep} implements a four-stage evaluation pipeline:
  \begin{enumerate}[leftmargin=1.4em]
      \item \textbf{Agent Output Scoring}: Individual agent responses are scored on a 0-100 scale using task-specific rubrics.
      \item \textbf{Binary Success Evaluation}: A calibrated LLM judge determines whether the merged output satisfies task requirements, yielding $E \in \{\text{\textsc{success}}, \text{\textsc{failure}}\}$.
      \item \textbf{Completeness Assessment}: The judge evaluates whether additional agent input would improve the response quality.
      \item \textbf{Agent Refinement}: Underperforming agent outputs are iteratively improved based on judge critiques.
  \end{enumerate}
  Judge calibration is performed on a held-out validation set of 200 labeled examples to estimate false positive/negative rates and set decision thresholds.

  \subsubsection{Agent Zoo Specification}
  The experimental agent population consists of two classes:
  \begin{itemize}[leftmargin=1.4em]
      \item \textbf{RAG Agents}: Domain-specific agents (\texttt{ExpertAgent}) with retrieval augmentation over curated knowledge bases. Domains include mathematics, law, finance, biology, medicine, and electrical engineering. Each agent loads domain-specific datasets and uses
  specialized prompt templates with retrieval-grounded context.
      \item \textbf{Conversational Agents}: LLM-based agents (\texttt{ConversationalAgent}) without retrieval, covering fitness, literature, technology, geography, storytelling, politics, academics, career guidance, trivia, and daily planning. Agent configurations are specified in YAML
   format with domain-specific constraints and prompt templates.
  \end{itemize}
  All agents are initialized with uniform priors $\alpha_0 = \beta_0 = 1$ unless memory-aware initialization is enabled.

  \subsubsection{Information Merging and Trust Weighting}
  The \texttt{InfoMergeStep} aggregates agent responses using trust-weighted selection:
  \begin{enumerate}[leftmargin=1.4em]
      \item Compute trust scores $t_i = \alpha_i / (\alpha_i + \beta_i)$ for each contributing agent.
      \item Filter responses from agents marked as "bad" (belief score below threshold).
      \item Merge remaining responses using LLM-based synthesis weighted by trust scores.
      \item Validate merged output through evidence-grounding and consistency checks.
  \end{enumerate}

  \subsubsection{Experimental Infrastructure}
  The evaluation framework (\texttt{run\_belief\_experiment.py}) supports:
  \begin{itemize}[leftmargin=1.4em]
      \item \textbf{Configurable Agent Selection}: Systematic sampling from the agent zoo with controllable population size.
      \item \textbf{Question Processing Pipeline}: Batch processing with configurable delays and timeout handling.
      \item \textbf{Comprehensive Logging}: Results are logged to structured JSON files including initial outputs, recursive delegation traces, final merged responses, and detailed performance metrics.
      \item \textbf{Statistical Validation}: Built-in A/B testing, bootstrap confidence intervals, and performance benchmarking capabilities.
  \end{itemize}

  \subsubsection{Evaluation Metrics Implementation}
  Quality assessment employs a multi-faceted scoring system:
  \begin{itemize}[leftmargin=1.4em]
      \item \textbf{Output Quality}: 0-100 scale scoring of initial vs. merged outputs using task-specific rubrics.
      \item \textbf{Quality Gains}: Absolute improvement (merged - initial) and relative improvement ((merged - initial) / initial).
      \item \textbf{Agent Contribution Tracking}: Classification of agents as "contributing" (positive impact) vs. "bad" (negative impact) based on comparative evaluation.
      \item \textbf{Routing Statistics}: Delegation depth, agent selection frequency, and belief evolution trajectories.
  \end{itemize}

  \subsubsection{Reproducibility and Configuration}
  All experiments are reproducible through:
  \begin{itemize}[leftmargin=1.4em]
      \item \textbf{Deterministic Sampling}: Fixed random seeds for Thompson sampling and LLM generation.
      \item \textbf{Configuration Management}: YAML-based agent specifications and experimental parameters.
      \item \textbf{Version Control}: Git-tracked experimental runs with commit hashes logged in results.
      \item \textbf{Environment Specification}: Docker containers with fixed dependency versions and Azure OpenAI API configurations.
  \end{itemize}

  \subsubsection{Computational Requirements}
  Typical experimental runs require:
  \begin{itemize}[leftmargin=1.4em]
      \item \textbf{Hardware}: 16GB RAM, 4-core CPU for coordination logic; GPU optional for local LLM inference.
      \item \textbf{API Costs}: \$0.50-2.00 per task depending on recursion depth and agent complexity.
      \item \textbf{Runtime}: 2-5 minutes per task with Azure OpenAI; 30-60 seconds with local models.
      \item \textbf{Storage}: 10-50MB per 100 tasks for complete logs and belief persistence.
  \end{itemize}

  \subsection{LLM-in-the-Loop Evaluation}
  Our pipeline uses LLMs for task generation, agent execution, and judging, which raises a potential circularity concern. We mitigate this in three ways:
  (i)~programmatic metrics (EM, F1, unit-test pass rates) short-circuit the judge for unambiguous cases, grounding a substantial fraction of verdicts in non-LLM signals;
  (ii)~the judge is calibrated on 200 human-labeled examples with measured FP/FN rates, so its error profile is quantified rather than assumed reliable; and
  (iii)~agent outputs and judge verdicts are produced by independent LLM calls with no shared state, preventing self-reinforcing feedback loops.
  The routing controller itself is purely probabilistic (Beta-Bernoulli updates) and contains no learned LLM parameters.
  While fully human-grounded evaluation remains the gold standard, these safeguards ensure that the LLM-based components are calibrated, partially redundant with programmatic checks, and structurally decoupled.

  \subsection{Proof of Theorem~\ref{thm:noisy-regret}: Regret Under Noisy Judge Feedback}
  \label{app:noisy-regret-proof}

  \paragraph{Setup.}
  There are $N$ arms (agents) with true Bernoulli success probabilities
  $\theta_1,\dots,\theta_N$. The judge observes each outcome through a
  binary symmetric--like channel with false-positive rate
  $\varepsilon_{\mathrm{FP}}$ and false-negative rate $\varepsilon_{\mathrm{FN}}$,
  so the observed success probability of arm $i$ is
  \[
    \tilde{\theta}_i
    \;=\; (1-\varepsilon_{\mathrm{FN}})\,\theta_i + \varepsilon_{\mathrm{FP}}\,(1-\theta_i)
    \;=\; \delta\,\theta_i + \varepsilon_{\mathrm{FP}},
  \]
  where $\delta = 1 - \varepsilon_{\mathrm{FP}} - \varepsilon_{\mathrm{FN}} > 0$.

  \begin{lemma}[Order preservation]\label{lem:order}
  Since $\delta>0$, the map $\theta_i\mapsto\tilde{\theta}_i$ is strictly
  increasing. Hence $\theta_{i^*}>\theta_j$ if and only if
  $\tilde{\theta}_{i^*}>\tilde{\theta}_j$; the optimal arm is unchanged.
  \end{lemma}

  \begin{lemma}[Gap contraction]\label{lem:gap}
  For any suboptimal arm $j$, the observed gap satisfies
  $\tilde{\Delta}_j = \tilde{\theta}_{i^*} - \tilde{\theta}_j = \delta\,\Delta_j$,
  where $\Delta_j = \theta_{i^*} - \theta_j$.
  \end{lemma}

  \paragraph{Main argument.}
  The controller never sees true outcomes; it runs Thompson sampling on the
  \emph{observed} Bernoulli problem $\{\tilde{\theta}_i\}_{i=1}^N$.
  By Lemma~\ref{lem:order}, the optimal observed arm coincides with the truly
  optimal arm $i^*$.
  Applying the Bayesian regret bound of \citet{AgrawalGoyal13}
  (Theorem~2 therein) to the observed problem yields
  \[
    \widetilde{R}(T)
    \;=\; \sum_{j\neq i^*} \tilde{\Delta}_j\,\mathbb{E}[n_j(T)]
    \;=\; O\!\bigl(\sqrt{NT\log T}\bigr).
  \]
  The true regret decomposes as
  \begin{align}
    R(T)
    &= \sum_{t=1}^{T}\bigl(\theta_{i^*}-\theta_{I_t}\bigr)
     = \sum_{j\neq i^*}\Delta_j\,\mathbb{E}[n_j(T)] \notag \\
    &= \sum_{j\neq i^*}\frac{\tilde{\Delta}_j}{\delta}\,\mathbb{E}[n_j(T)]
     = \frac{1}{\delta}\,\widetilde{R}(T)
     = O\!\Bigl(\frac{\sqrt{NT\log T}}{\delta}\Bigr),
  \end{align}
  where we used Lemma~\ref{lem:gap} in the penultimate step.
  This completes the proof.\qed

\end{document}